# Unsupervised Active Learning in Large Domains

**Harald Steck**
Artificial Intelligence Laboratory
Massachusetts Institute of Technology
Cambridge, MA 02139
*harald@ai.mit.edu*

**Tommi S. Jaakkola**
Artificial Intelligence Laboratory
Massachusetts Institute of Technology
Cambridge, MA 02139
*tommi@ai.mit.edu*

**Abstract**

Active learning is a powerful approach to analyzing data effectively. We show that the feasibility of active learning depends crucially on the choice of measure with respect to which the query is being optimized. The standard information gain, for example, does not permit an accurate evaluation with a small committee, a representative subset of the model space. We propose a surrogate measure requiring only a small committee and discuss the properties of this new measure. We devise, in addition, a bootstrap approach for committee selection. The advantages of this approach are illustrated in the context of recovering (regulatory) network models.

## 1 INTRODUCTION

Many domains of interest involve a large collection of inter-dependent random variables $X$, where the associated models have to be inferred under severe sampling constraints. Ensuring reliable predictions of outcomes of perturbations or deliberate interventions is challenging. This is precisely the setting in functional genomics, for example, where models of genetic regulatory networks need to be estimated on the basis of a very limited number and type of observations about the variables (genes). It is frequently more informative to intervene in the system to elicit a response rather than to rely on purely observational data. This is indeed necessary to be able to predict outcomes of interventions. In functional genomics context, the interventions might mean "knocking out" or "over-expressing" a gene (rarely multiple genes).

The overall problem can be formalized as follows. We set certain variables $Q \subset X$ to their desired value $q$ and observe a sample from the rest. Active learning in this setting seeks to determine the intervention which can be expected to yield the most informative response from the system. This expectation depends on the current knowledge of the active learner, often represented by a committee, a representative subset of the model space used. The objective is to minimize the overall cost of acquiring the data (e.g., the number of queries) necessary to ensure reliable predictions.[1]

Various approaches to active learning have been developed, mostly for supervised learning and regression. Relatively few approaches have addressed *unsupervised* active learning in the above sense; the most recent ones include [14, 15, 10]. While relevant, such approaches typically require enormous computational resources, limiting their applicability to domains with a fairly small number of variables.

In this paper we develop an approach to unsupervised active learning which is tailored to large domains. We show that the computational efficiency depends crucially on the properties of the measure with respect to which the optimal query is chosen, additivity in particular (cf. Section 5). The standard information gain turns out to require a committee size that scales exponentially with the size of the domain (the number of random variables). We propose a new measure, which we term "average KL divergence of pairs" (KL2). It emerges as a natural extension to the query by committee approach [12, 4]. In Section 3, we give a general motivation for this approach, and generalize it to *un*supervised active learning. This is followed by a bootstrap approach for selecting the committee members (cf. Section 6) and an empirical evaluation (cf. Section 9).

## 2 ACTIVE LEARNING

Before we develop an approach to unsupervised active learning, let us first formalize the basic idea of interventional learning, where the learner can impose values on a subset of the variables and observe the response. We denote an intervention by $do(q)$ [11]. The observed instantiation of the remaining variables $X \setminus Q$ is then drawn from the distribution $P(X|\text{do}(q))$.[2] Note that the distribution $P(X|\text{do}(q))$ may differ from the conditional distribution $P(X|q)$, as the query variables are *actively* set to some value by the learner. For instance, when a Bayesian network model is used to describe the distribution over $X$, the distribution $P(X|\text{do}(q))$ is given by the *mutilated* network, which is obtained by setting $Q = q$ *and* removing all the directed edges pointing *towards* the variables in $Q$ [11]. For brevity, we will loosely denote an intervention by $q$ in the remainder of the paper.

---
[1] We include the model structure here as a "prediction".

A common assumption in all approaches to active learning is that the model class used by the active learner is actually capable of describing the true distribution underlying the data. Concerning Bayesian networks, this corresponds to the Causal Markov assumption and the Faithfulness assumption [13]. Moreover, we assume that there are no variables unknown to the learner, and that the variables are discrete.

## 3 QUERY BY COMMITTEE

The *Query by Committee* approach uses a (small) committee of experts, which represents the current knowledge of the active learner. The original formulation is for supervised learning, assuming no noise concerning the labels [12, 4]. The basic idea is to query a missing label if there is a high degree of disagreement among predictions of the committee. In active learning of a binary classifier, for example, the disagreement among a committee is maximal if half the experts predict the one label, while the other half does the opposite. The degree of disagreement serves as an estimate for the (myopic) information gain in this approach. While maximizing the expected instantaneous information gain does not necessarily optimize prediction accuracy, the two objectives co-vary for several important classes of models [4].

Query by committee can be motivated from first principles of information theory: The information obtained from an experiment *after* observing its outcome is equivalent to the uncertainty *prior* to the observation of its outcome. The degree of disagreement of the committee can serve as an estimate of this uncertainty. In contrast to supervised learning, in unsupervised learning we have to define a measure of variability to quantify the degree of disagreement among the committee members. In the following, we examine several possible measures besides the standard measure of information gain, which is the only consistent such measure in the sense of Shannon entropy.

---

[2]We loosely use $P(X|\text{do}(q))$ in order to denote the distribution over $X \setminus Q$.

## 4 MEASURES OF DISAGREEMENT

In *un*supervised active learning, each model $m$ in the committee $C$ describes a distribution over $X$. This suggests to quantify the disagreement among any two committee members $m_1, m_2 \in C$ by a distance or divergence measure $D^{(2)}$, where $D^{(2)}$ is applied to the associated probability distributions, $D^{(2)}(P(X|q, m_1) || P(X|q, m_2))$.[3] To capture the *overall* disagreement among the committee members, we impose the following desiderata for the measure $D$:

1. *symmetry*: $D$ depends only on the unordered *set* of experts

2. *weights*: $D$ admits different weights such as $P(m_i)$ assigned to the various experts $m_i$

3. *additivity*: the measure of disagreement $D$ in a domain is the *sum* of the measures in *independent* subdomains.

Based on any pairwise divergence measure $D^{(2)}$, one can obtain generalized measures complying with the first two properties by taking weighted (arithmetic) averages, denoted by $\langle \cdot \rangle_{m \sim P(M)}$. As there are three possible ways of averaging, this gives rise to three different generalized measures concerning a committee $C$ of $K = |C|$ experts:

- $\left\langle D^{(2)}\left(P(X|q,m) \middle|\middle| \langle P(X|q,m') \rangle_{m' \sim P(M)}\right)\right\rangle_{m \sim P(M)}$

- $\left\langle D^{(2)}\left(\langle P(X|q,m) \rangle_{m \sim P(M)} \middle|\middle| P(X|q,m')\right)\right\rangle_{m' \sim P(M)}$

- $\left\langle \left\langle D^{(2)}(P(X|q,m) || P(X|q,m')) \right\rangle_{m \sim P(M)} \right\rangle_{m' \sim P(M)}$

These three generalized measures behave in different ways concerning the additivity. This is outlined in the next section, where we also discuss lower bounds on the committee size.

## 5 ADDITIVITY PROPERTY AND COMMITTEE SIZE

The additivity property of the measure ensures that it "scales" in the right way with the size of the domain. This is an important property of measures, see, e.g., [8]. If a domain $X$ can be decomposed into *independent* subdomains $X_i$ ($X = \bigcup_i X_i$) such that $P(X) = \prod_i P(X_i)$, and given that the model class $M$ under consideration is capable of representing these independences, i.e., $P(X|M) = \prod_i P(X_i|M)$, then an

---

[3]For brevity of notation, we write $P(\cdot|\cdot)$ instead of $P(\cdot|\cdot, \mathcal{D})$ in the remainder of this paper, as $\mathcal{D}$ appears in every term.

*additive* measure $D$ decomposes accordingly, namely like $D(P(X|M)) = \sum_i D(P(X_i|M))$.

The additivity property of a measure is also crucial for feasible computations in practice. This is primarily for three reasons. First, in large domains, the measure $D(X|q)$ has to be evaluated approximately, as it is typically intractable to sum explicitly over all configurations of $X$. It is hence desirable that the scores $D(..|q)$ and $D(..|q')$ of queries $q$ and $q'$ differ by a large margin: a large margin permits large approximation errors while it is still possible to identify the query with the largest score. By allowing more errors, faster evaluations are possible. Moreover, the additivity of the measure ensures that the margin does not decrease as the size of the domain increases. In contrast, the margin of a subadditive measure diminishes with increasing domain size. Finally, as explained below, additivity determines the minimal size of the committee. A measure that retains additivity for small committees is computationally more efficient.

### 5.1 PAIRWISE MEASURE

The additivity of the pairwise divergence measure $D^{(2)}(P||Q)$ is a necessary requirement for the additivity of the measures concerning $K$ experts. This suggests that we use the Kullback-Leibler (KL) divergence, as $D^{KL}(P(X)||Q(X)) = \sum_i D^{KL}(P(X_i)||Q(X_i))$ if $P(X) = \prod_i P(X_i)$ and $Q(X) = \prod_i Q(X_i)$.

### 5.2 GENERALIZED MEASURES

We focus here on the generalized measures that employ the KL divergence as the pairwise measure. Then the three generalized measures introduced in Section 4 become the Jensen-Shannon divergence (JS), the backward JS divergence (BJS), and the measure we call "average KL divergence of pairs" (KL2).

#### 5.2.1 Jensen-Shannon Divergence

The JS divergence is equivalent to the information gain, and relates to the entropy in a standard way:

$$D^{JS}_{m \sim P(M)}(P(X|q,m)) \quad (1)$$
$$\equiv \left\langle D^{KL}\left(P(X|q,m) \middle\| \langle P(X|q,m')\rangle_{P(M)}\right)\right\rangle_{P(M)}$$
$$= H(P(X|q)) - \left\langle H(P(X|q,m))\right\rangle_{P(M)}$$

If the Bayesian rule

$$P(X|q,M) = \frac{P(X|q)}{P(M)}P(M|X,q) \quad (2)$$

holds,[4] the JS divergence can also be understood in terms of the distributions over the model space $M$ before and after the next observation of $X$ given the query $q$:

$$D^{JS}_{m \sim P(M)}(P(X|q,m))$$
$$= \left\langle D^{KL}\left(P(M|x,q)\middle\|P(M)\right)\right\rangle_{P(X|q)}$$
$$= H(P(M)) - \left\langle H(P(M|x,q))\right\rangle_{P(X|q)}$$

Thus maximizing the disagreement among the committee, when measured in terms of the JS divergence, is equivalent to minimizing the expected posterior entropy of the search space. The latter approach essentially was taken in [10, 14]. The entropy can be understood as a measure of the effective size of the search space given the data. Hence, maximizing the disagreement corresponds to maximally shrinking the effective size of the search space.

Another well-known property of the JS divergence is that it is bounded from above. This upper bound depends on the number of experts. Given a fixed committee size, JS divergence cannot be additive for arbitrarily large domains. Indeed, even when each of the experts $m$ represents the correct independences, say, $P(X|q,m) = \prod_i P(X_i|q,m)$, JS is not necessarily additive. It is additive only if the weights assigned to the committee members also factor accordingly:

$$P(M) = \prod_i P(M_i), \quad (3)$$

where $M_i$ are the sets of subgraphs pertaining to the subdomain $X_i$, i.e., $P(X_i|M) = P(X_i|M_i)$. Put another way, for JS divergence to be additive $\left\langle \prod_i P(X_i|q,m)\right\rangle_{m \sim P(M)} = \prod_i \left\langle P(X_i|q,m)\right\rangle_{m \sim P(M)}$ must hold.

The fact that $P(M)$ has to factor according to Eq. 3 leads to a (pessimistic) lower bound on the committee size $|C|$. Given $L$ independent subdomains, JS divergence can be additive only if

$$|C^{JS}| \geq \prod_{i=1}^{L} |M_i| \geq 2^L \quad (4)$$

where $|M_i|$ denotes the number of alternative subgraphs due to model uncertainty. The committee size therefore has to grow exponentially with the number of independent subdomains involving uncertainty. Any feasible committee size is typically much too small to achieve (full) additivity of the JS divergence.

---

[4]The Bayesian rule holds, of course, when the distributions are *conditioned* on the query $q$. However, when $q$ is replaced by the intervention $do(q)$, the Bayesian rule need not hold in general. Concerning Bayesian networks, Eq. 2 holds for interventions $do(q)$ given parameter modularity and parameter independence [15]. Also note that $P(M|q) = P(M)$.

### 5.2.2 Backward JS Divergence

The backward JS divergence (BJS) is very similar to the JS divergence. For BJS divergence the two arguments in the KL divergence are swapped,

$$
\begin{aligned}
& D^{\text{BJS}}_{m \sim P(M)}(P(X|q,m)) \quad (5) \\
& \equiv \left\langle D^{\text{KL}}\left(\langle P(X|q,m)\rangle_{P(M)} \middle\| P(X|q,m')\right)\right\rangle_{P(M)} \\
& = D^{\text{KL}}\left(\langle P(X|q,m)\rangle_{P(M)} \middle\| \langle P(X|q,m')\rangle^{\text{geom}}_{P(M)}\right) \\
& = \left\langle D^{\text{KL}}\left(P(M) \middle\| P(M|x,q)\right)\right\rangle_{P(X|q)}
\end{aligned}
$$

illustrating that the similarity extends to model space. While BJS divergence is a measure of variance, it does not measure information gain. Note that the above derivation shows that BJS divergence can be viewed as a KL divergence between the weighted *arithmetic* average , $\langle P(X|m)\rangle_{P(M)}$, and the weighted *geometric* average, $\langle P(X|m)\rangle^{\text{geom}}_{P(M)} = \prod_m P(X|m)^{P(m)}$ (where we use KL divergence for unnormalized measures).

Like JS divergence, BJS divergence can be additive only if the committee size exceeds the lower bound given in Eq. 4. This follows essentially from the arithmetic averages taken inside the logarithm.

### 5.2.3 KL2 Divergence

The third measure quantifies the disagreement in terms of the average KL divergence of all pairs of experts, a measure we refer to as KL2. When we have independent subdomains, i.e., $P(X) = \prod_i P(X_i)$, KL2 divergence is additive if each expert $m$ on the committee represents this independence: $P(X|m) = \prod_i P(X_i|m)$. In contrast to the previous measures, there is no constraint concerning the distribution $P(M)$ over experts. Consequently, the committee only has to be large enough so that it can represent the model uncertainty in each of the different subdomains, independently of each other. In particular, the overall uncertainty in the domain need not be represented in the committee. A lower bound on the committee size is hence given by the subdomain with the largest uncertainty:

$$|C^{\text{KL2}}| \geq \max_i |M_i| \geq 2. \quad (6)$$

The minimal size of the committee is independent of the domain size. This leads to a desirable scaling of the committee size and can be expected to provide a computational advantage over the other two measures.

Note that, analogously to BJS, KL2 is a measure of variability, but it does not measure information gain. However, the following relation holds

$$
\begin{aligned}
& D^{\text{KL2}}_{m \sim P(M)}(P(X|q,m)) \quad (7) \\
& = D^{\text{JS}}_{m \sim P(M)}(P(X|q,m)) + D^{\text{BJS}}_{m \sim P(M)}(P(X|q,m))
\end{aligned}
$$

The derivation is given in the Appendix, together with additional properties of the KL2 divergence. Eq. 7 also implies that the KL2 divergence is an upper bound on the JS divergence, since all the measures are non-negative.

Besides the small committee size necessary for the additivity of the KL2 divergence, the fact that the KL2 divergence can be evaluated based on *pairs* of experts leads to efficient computations in practice. Since the number of pairwise evaluations grows quadratically with the number of experts, a small committee is particularly efficient.

## 6 CHOOSING A COMMITTEE

In the previous section, we gave lower bounds on the size of the committee, based on each measure of disagreement. If the class of models is tractable, e.g., in the case of a forest or trees [9], one may use the entire model space as the committee. In general, one can expect that the *stochasticity* of the learning process increases as the committee size shrinks. A disadvantage of increased stochasticity is that the progress in finding the unknown true model underlying the data may be decreased. On the other hand, an increased stochasticity concerning the queries may render the (myopic) learner more robust, as a larger part of the model space is explored. This may be particularly important at the beginning of the learning process when skewed initial data may easily confuse the learner.

There are two obvious ways of selecting the committee members:

- Markov Chain Monte Carlo (MCMC): find a committee based on the $K$ highest-scoring models [15, 10].

- bootstrap / bagging: The $K$ models comprising the committee are drawn *independently* of each other from the model space (with replacement), given the data $\mathcal{D}$ seen so far. This can be achieved by generating $K$ bootstrap samples $\mathcal{D}^i$ from the data $\mathcal{D}$, and by subsequently learning a model $m^i$ from each sample $\mathcal{D}^i$ ($i = 1, ..., K$).

We resort to the bootstrap approach as it can be used efficiently in finding a (small) committee that represents the search space. In the context of learning Bayesian networks, the bootstrap approach has been used successfully in [5] along with estimates of confidence intervals.

## 7 SEARCH FOR OPTIMAL QUERY

Once a committee has been determined, the aim of the (myopic) active learner is to find the query that maximizes the measure of disagreement. In large domains, however, it is typically intractable to compute the score of every possible query. This is because there are $2^N$ different subsets of (query) variables in a domain comprising $N$ variables, and each variable may have several states. Hence, some heuristic search strategy has to be devised in the space of queries. For computational efficiency, we use a *greedy* approach in this paper. This scheme proceeds in rounds of ascending number of query variables. In each round the highest-scoring variable (and its state) is added to the query. The greedy procedure terminates when the score cannot be increased by adding another variable to the current query. Moreover, the increase in each round is required to exceed a (small) threshold value. We devise this threshold value in order to account for noise in the scores, which are computed only approximately.

In large domains consisting of many small, nearly independent subdomains, the optimal query is approximately determined by the optimal queries pertaining to the subdomains. In this context the greedy strategy may actually find close-to-optimum queries.

## 8 BAYESIAN NETWORKS

In the remainder of this paper, we study active learning in Bayesian networks. We assume that the variables have a multinomial distribution, and treat the learning task as an optimization problem: we optimize the posterior probability of the network structure by local search, as described for observational data in [7]. In [3], this approach was extended to allow for both observational and experimental data.

Local search may introduce additional noise in the learning process. Its effect on the progress of learning, however, is hard to assess in large domains, since finding the optimal network is an NP-complete problem [2]. It is conceivable that the search strategy may be improved by enabling it to reuse various properties of previously learned networks(e.g., presence of edges). This is particularly reasonable when the posterior distribution over the model space does not change much between successive time steps of active learning. Such an improvement may not only lead to a reduction in the noise introduced by the search strategy, but may also considerably enhance the computational efficiency of the algorithm.

When active learning is carried out on *small* data sets, one can expect the resulting models to be *parsimonious* to avoid over-fitting. Sparsity of Bayesian network structures confers several advantages. First, a sparse graph facilitates an easy interpretation of the qualitative dependencies among the variables. Second, learning and inference are feasible. In particular, the computation of the KL2 divergence simplifies considerably, as for each pair of models $m$ and $m'$ the expression $D^{\text{KL}}(P(X|q,m)||P(X|q,m')) = \sum_j \sum_{x_j, \pi_j, \pi'_j} P(x_j, \pi_j, \pi'_j|q,m) \log \frac{P(x_j|\pi_j,q,m)}{P(x_j|\pi'_j,q,m')}$ can be computed efficiently (i.e., one can explicitly sum over all instantiations as the number of joint parents is small). Moreover, the marginals $P(x_j, \pi_j, \pi'_j|\text{do}(q), m)$ can be approximated efficiently by a small sample size obtained by forward-sampling (given the mutilated graph). Unfortunately, sparsity does not lead to any immediate simplifications of the JS divergence. In our experiments we evaluate the JS divergence approximately by importance sampling.

## 9 PRELIMINARY EXPERIMENTS

In our experiments, we compared the active learner employing KL2 and JS divergences to passive learning as well as to learning from randomly chosen interventions. To facilitate speedy evaluations, the committee included only *two* models. In our experiments here, a time step in active learning typically took about one minute on a 1.7 GHz P4 (with the simplest implementation). Since the computational effort for evaluating the KL2 measure increases quadratically with the committee size (cf. Section 5.2.3), large committee sizes may become prohibitive in large domains.

The objective was to recover the Bayesian network from which the data was sampled, given interventions. We carried out experiments with the widely-used alarm-network [1] and a highest-scoring regulatory network recovered from gene expression data along with other information sources (factor-gene binding information) [6]. The alarm network comprises 37 discrete variables with 2, 3, or 4 states and includes 46 edges. The regulatory network involves 33 variables, 56 edges, and each variable was discretized to 4 states. In our experiments with the alarm network, we carried out at most 200 learning steps. The regulatory

| algo | edge error | edge entr. | predictive accuracy: 0 | 1 | 5 | 10 |
|---|---|---|---|---|---|---|
| p | 74 | 199 | 1.5 | 2.4 | 6.0 | 7.9 |
| r: 2 var. | 59 | 183 | 1.1 | 1.3 | 1.9 | 2.4 |
| r: 5 var. | 52 | 167 | 1.2 | 1.2 | 1.4 | 1.5 |
| a: JS | 56 | 174 | 1.2 | 1.2 | 1.3 | 1.5 |
| a: KL2 | 46 | 153 | 1.3 | 1.2 | 1.1 | 1.1 |
| std | 2 | 6 | 0.1 | 0.2 | 0.2 | 0.2 |

Table 1: Alarm network experiment: We assessed the quality of the learned model structure and the predictive accuracy given 0, 2, 5 and 10 interventions. The models were learned in different ways: (p) passive, (r) random queries of 2 or 5 variables and (a) active learning using JS and KL2 divergences. All values were obtained by averaging over 5 trials; the bottom row shows the standard deviation.

| algo | q.v. | edg. err. | edg. entr. | predictive accuracy: 0 | 1 | 5 | 10 |
|---|---|---|---|---|---|---|---|
| p |  | 37 | 65 | 1.8 | 2.4 | 3.8 | 4.2 |
| r | 1 | 30 | 54 | 1.7 | 1.9 | 2.5 | 2.7 |
| r | 5 | 21 | 33 | 1.5 | 1.6 | 1.7 | 1.7 |
| a: JS | $\leq 1$ | 24 | 43 | 1.4 | 1.5 | 1.8 | 2.0 |
| a: KL2 | $\leq 1$ | 27 | 42 | 1.6 | 1.6 | 1.8 | 1.9 |
| a: JS |  | 21 | 37 | 1.4 | 1.4 | 1.5 | 1.5 |
| a: KL2 |  | 16 | 26 | 1.6 | 1.6 | 1.5 | 1.3 |
| std |  | 1 | 2 | 0.2 | 0.2 | 0.2 | 0.2 |

Table 2: The quality of the models learned in the regulatory network experiment; regarding the abbreviations, see Table 1; the column (q.v.) indicates the restrictions on the number of query variables in the learning process.

network, originally derived from 320 expression measurements, involved up to 300 queries.

We assessed the learned network structures by the edge entropy and the edge error as defined in [15]; we computed these quantities based on a bootstrap approach of 200 subsamples (cf. also [5]). Note that the divergence measures employed by the active learners in this paper are actually not directly aimed at optimizing these edge-wise quantities. We also assessed the predictive accuracy of the learned models, similarly to [10]. Besides the ability to predict the marginal distribution over the variables, we also assessed a model's ability to predict the effects of interventions. We quantified this in terms of the KL divergence between the original model and the learned model. We averaged the KL divergence over all possible interventions involving one or two variables, as well as over 100 randomly chosen interventions on 5 and 10 variables, respectively.

The results are depicted in the Tables 1 and 2. It is obvious that the edge error as well as the edge entropy could be reduced considerably by active learning. Moreover, in large networks the *number* of variables queried simultaneously appears to have a crucial impact on the quality of the learned models. This applies not only to randomly chosen queries, but also to the active learner: in the alarm network experiments, the active learner employing the JS divergence intervened on 1.6 variables on average over the 200 time steps, while KL2 divergence involved 9.2 query variables on average. The average numbers of query variables in the regulatory network experiments were 2.1 and 7.2, respectively. This difference can be explained in terms of the additivity property (cf. Section 5): even with only two committee members, KL2 divergence can be expected to be "nearly" additive when the *sparse* networks contain several "nearly" independent components. In contrast, the small committee size causes the JS divergence to be subadditive in our experiments, leading to relatively *small* margins between the scores of different queries (cf. Section 5). Due to the relatively large number of query variables, the active learner employing the KL2 divergence yields the smallest edge error and edge entropy in both experiments.

We also examined the reliability of individual edges in the recovered regulatory network. Again, we took a bootstrap approach based on 200 subsamples. The active learner employing KL2 divergence resulted in 31 edges with $P > 90\%$, another 13 edges with $P > 50\%$ and additional 4 edges with $P > 30\%$. All these 48 edges are indeed present in the original network structure. We examined the 8 edges that were erroneously missing in the learned networks ($P < 10\%$), and found that their absence is favored by the scoring function (posterior probability) on data sets of size 300; this is because of the penalty for model complexity inherent in the posterior probability, which helps avoid overfitting when the training set is small.

Besides the graph structure, active learning obviously also improves the predictive accuracy compared to passive learning, in particular when it comes to predicting the effects of *several* interventions. This also applies to learning from randomly chosen queries: if the number of query variables is increased (not exceeding some optimal value), the predictive accuracy also improves. The results suggest that, compared to JS divergence, KL2 favors trading some predictive accuracy from a *few* simultaneous interventions for an improved accuracy concerning *several* interventions. We found that this behavior becomes more pronounced with smaller data sets, e.g., after 100 time steps. These differences are not significant, however. Moreover, since KL2 yields low errors concerning the model structure, this suggests that the "right" network structure is particularly important for prediction the effects of *numerous* interventions. As KL2 divergence achieves this low error in model structure by querying several variables during the learning process, there is less data available for parameter estimation. This may explain the slightly increased prediction error concerning a *small* number of interventions, where the model structure seems to be less important.

In addition to the number of query variables, it also matters which of the variables and which of their states are chosen. In our experiments, picking 5 query variables randomly leads to about the same results as the active learner employing JS divergence, although the latter queries considerably fewer variables.

As it is rare in biological experiments to be able to intervene on numerous genes simultaneously, we also studied the active learner with queries restricted to at most one variable. In this case we can easily find the optimal query with respect to the JS divergence since (sub)additivity is relevant only with multiple-variable queries. It is not surprising that in this context the JS divergence, as a valid measure of information gain, performs slightly better than the surrogate KL2 measure. The quality of the learned models along with the relevance of KL2 could be improved by intervening on several genes simultaneously.

## 10 CONCLUSIONS

In this paper we have generalized the *Query by Committee* approach to unsupervised setting with special attention to computational efficiency. We examined three divergence measures to quantify the degree of disagreement among the committee members and demonstrated that additivity of the measure is crucial for feasible computations, particularly in terms of the committee size. The standard information gain (JS divergence) requires a committee size that grows exponentially with the size of the domain, while our surrogate KL2 measure remains additive – and therefore can be evaluated efficiently – even when based on a small committee. Apart from that, we showed that a bootstrap approach can serve as an efficient tool for selecting a small committee, as the members are drawn independently from the model space given the data.

A number of extensions are possible. For example, our experiments involve only a very limited exploration of the committee size and its effect on the quality of the queries. We are also developing stronger theoretical guarantees for additive measures.

## Acknowledgements

We would like to thank Alexander Hartemink for providing us with the highest-scoring Bayesian network obtained from gene expression data [6], and the anonymous reviewers for valuable comments. Harald Steck acknowledges support from the German Research Foundation (DFG) under grant STE 1045/1-1. Tommi Jaakkola acknowledges supported from Nippon Telegraph and Telephone Corporation and from NSF ITR grant IIS-0085836.

# APPENDIX

The KL2 divergence can be rewritten in various ways, including:

$$D^{\text{KL2}}_{m\sim P(M)}(P(X|m)) \tag{8}$$
$$\equiv \left\langle \left\langle D_{\text{KL}}\left(P(X|m)\middle\|P(X|m')\right)\right\rangle_{P(M)}\right\rangle_{P(M)}$$
$$= \left\langle D^{\text{KL}}\left(P(X|q,m)\middle\|\langle P(X|q,m')\rangle^{\text{geom}}_{P(M)}\right)\right\rangle_{P(M)}$$

The last line is obtained similarly to BJS divergence (cf. Section 5.2.2). It shows that KL2 divergence can be viewed as a measure of the average disagreement between each of the experts and their weighted *geometric* average. Qualitatively, this differs from JS divergence only by replacing an *arithmetic* average with a *geometric* average. Based on the above derivation, Eq. 7 in the text now follows from summing up Eqs. 1 and 5, where the arithmetic averages inside of the KL divergences cancel out.

To make the discussion concerning KL2 divergence more complete, let us briefly mention another way of rewriting this measure:

$$D^{\text{KL2}}_{m\sim P(M)}(P(X|m)) \tag{9}$$
$$= \left\langle \sum_x \left[P(x|m) - \left\langle P(x|m')\right\rangle_{P(M)}\right] \cdot \log P(x|m)\right\rangle_{P(M)}$$
$$= \left\langle \sum_x \left[P(x|m) - \langle P(x|m')\rangle_{P(M)}\right] \cdot \log \frac{P(x|m)}{\langle P(x|m')\rangle_{P(M)}}\right\rangle_{P(M)}$$

The second line shows that, in contrast to the JS divergence, here the difference between each of the experts and the arithmetic average is taken concerning the *weights* of the log-term, $P(x|m)$, rather than with respect to the log-terms. This is obtained by first expanding the expression which defines KL2 divergence, and then renaming the indices. The last line in Eq. 9 follows immediately from the second line, and illustrates again that KL2 is the sum of JS and BJS (cf. Eq. 7 in the text).